\documentclass[runningheads]{llncs}
\usepackage[T1]{fontenc}
\usepackage{graphicx}
\usepackage{booktabs}
\usepackage{amsmath}
\usepackage{siunitx}
\usepackage{multirow}
\usepackage{color}
\usepackage{url}

\usepackage{hyperref}
\begin{document}
\title{Data-Driven Glyph Positioning for Handwritten Music Synthesis}
%
%
\author{Elona Shatri\inst{1}\orcidID{0000-0002-1651-5848} \and
Ji\v{r}\'{i} Mayer\inst{1}\orcidID{0000-0001-6503-3442} \and
Jan Haji\v{c} jr.\inst{1}\orcidID{0000-0002-9207-567X}}
\authorrunning{E. Shatri, J. Mayer, J. Haji\v{c} jr.}
\institute{Charles University, Prague, Czech Republic\\
\email{\{shatri,mayer,hajic\}@ufal.mff.cuni.cz}}
\maketitle

\begin{abstract}

Synthesised data is a common strategy to address the lack of annotated handwritten music scores in Optical Music Recognition (OMR). Smashcima is a recent synthesis framework that assembles pages of handwritten scores by sampling glyph images from MUSCIMA++ and positioning them with hard-coded rules. In this work we ask whether these rules are grounded empirically, and whether a learned model for context-dependent positioning can further improve synthesis quality. We extract inter-glyph distributions from ${\sim}$285 thousand annotated relationships across MUSCIMA++ and OmniOMR, finding that several key parameters deviate systematically from real handwriting. Replacing these constants with empirical means reduces positioning RMSE by up to 60\%. We then train a Graph Neural Network (GNN) on the Music Notation Graph structure to predict context-dependent offsets, achieving an additional 50--85\% RMSE reduction on stems and beams by modelling how neighbouring glyphs constrain placement. Cross-dataset analysis shows that spacing differences between the two corpora are explained by note density (3.3$\times$), not style, and that duration-proportional spacing does not hold in handwriting ($\text{R}^2 = 0.001$). In downstream evaluation, a YOLOv8 detector pretrained on GNN-positioned synthetic data achieves +30\% mAP50 over the baseline and +11\% over empirical means when fine-tuned on 200 real handwritten staves, showing that positioning accuracy matters for downstream OMR.

\end{abstract}

\section{Introduction}
\label{sec:intro}

An OMR system's performance highly depends on the availability of training data. Manually annotating handwritten scores is labour-intensive and there are only a few relatively small such datasets for the purpose of training deep learning models. To address this there are synthesis frameworks which generate realistic training images from symbolic music representations. However, recent work has identified a ``glass ceiling'' on the effectiveness of synthetic training data~\cite{alfaro2023optical}, where increasing the volume of synthetic data yields diminishing returns. Understanding what this ceiling is made of and how to get past it requires decomposing the synthetic-to-real domain gap into its main factors such as symbol appearance, paper texture, and spatial layout.

In this paper, we focus on the layout component.
Smashcima~\cite{smashcima} is an open-source synthesis tool that takes a MusicXML file and produces rastered images of handwritten music together with pixel-level annotations by sampling real glyph images from MUSCIMA++~\cite{muscimapp} and positioning them according to a set of rules. However, the numerical parameters of these rules (stem length, accidental offset, beam angle, inter-note spacing) are constants chosen by visual inspection rather than measured from data. Moreover, these rules treat each relationship independently, they cannot model how the position of one glyph constrains its neighbours.

In this work we address two questions: (1) how far do these constants deviate from real handwriting distributions, and (2) can a learned model that captures context-dependent positioning further improve synthesis quality? We make the following contributions:

\begin{enumerate}
    \item A systematic empirical characterisation of inter-glyph positioning distributions across 199 relationship types, extracted from two complementary annotated corpora (${\sim}$285k relationships).
    \item A Graph Neural Network that predicts context-dependent glyph offsets from the Music Notation Graph structure, reducing positioning RMSE by 50--85\% over empirical means on stems and beams.
    \item Cross-dataset validation showing that spacing differences between controlled and real-world corpora are primarily a density effect (3.3$\times$), not a style difference, and that duration-proportional spacing does not hold in handwriting ($\text{R}^2 = 0.001$).
    \item A downstream evaluation showing that GNN-positioned synthesis improves detection mAP50 by +30\% over the baseline and +11\% over empirical means on real handwritten staves with 200-shot fine-tuning.
\end{enumerate}

\section{Related Work}
\label{sec:related}




Synthetic data for OMR training has been generated using engraving engines such as Verovio~\cite{calvo2018camera}, LilyPond~\cite{tuggener2018deepscores}, Dorico~\cite{Shatri_tenor2020} producing typeset scores at scale. However, these do not model handwriting variation. For handwritten synthesis, GANs were used to generate realistic music symbol images for data augmentation~\cite{shatri2024synthesising,hande2023towards,tirupati2024crafting}, addressing the appearance component of the gap between synthetic and real. Gasbert et al.~\cite{gasbert2024} combine GAN-generated glyphs with Smashcima's layout engine. We address the layout component by improving how symbols are positioned relative to each other.

Smashcima~\cite{smashcima} samples glyph images from MUSCIMA++ and assembles them into full pages using a column-based engine. It is the only framework that produces handwritten notation with pixel-level annotations from MusicXML input. However, its positioning rules use hard-coded constants that have not been validated against data.

MUSCIMA++~\cite{muscimapp} provides 91,255 annotated symbols across 140 pages from 50 writers, with 82,261 explicit relationships encoded as a directed graph (Music Notation Graph, MuNG). OmniOMR~\cite{omniomr} is a more recent dataset containing real-world handwritten scores with MuNG annotations.

Automatic score engraving has been addressed with graph neural networks. Foscarin et al.~\cite{foscarin2024} proposed a GNN for voice and staff prediction, and subsequent work~\cite{engravinggnn} extended this to multi-task engraving decisions including stem direction and clef assignment, using architectures from GraphMuse~\cite{karystinaios2024graphmuse}. These systems solve the semantic engraving problem (what notation to use), while our work addresses the visual positioning problem: where to place glyphs in millimetre space with realistic handwriting variation.

\section{Data and Method}
\label{sec:method}

\textbf{MUSCIMA++} uses the MuNG annotation which provides directed edges between glyphs encoding attachment relationships (e.g., notehead$\to$stem, notehead $ \to $ accidental). We extract 143,539 directed edges across 199 unique relationship types.

\textbf{OmniOMR} contains 100 page-level documents from diverse real-world handwritten scores, annotated in MuNG format. Unlike MUSCIMA++, documents represent different musical pieces by different writers under natural writing conditions. We extract 141,716 directed edges.

\subsection{Feature Extraction}

We compute the relative offset $(dx, dy)$ in millimetres between the bounding box centres of the source and target glyphs for each directed edge in the MuNG graph:
\begin{equation}
    dx = \frac{cx_{\text{tgt}} - cx_{\text{src}}}{\text{DPI}} \times 25.4, \quad
    dy = \frac{cy_{\text{tgt}} - cy_{\text{src}}}{\text{DPI}} \times 25.4
\end{equation}
where DPI is determined per document (MUSCIMA++: 300; OmniOMR: 185 to 290).

\subsection{Evaluation Protocol}

For evaluation we use leave-one-writer-out (LOWO) cross-validation on MUSCIMA++ and leave-one-document-out on OmniOMR. We compare three predictors: (1) Smashcima's hard-coded constant (baseline), (2) the empirical mean across all training writers (global model), and (3) a GNN that predicts offsets conditioned on the local graph neighbourhood. Performance is measured by RMSE in millimetres.

\section{How Far Off Are the Constants?}
\label{sec:results}

\begin{figure*}[t]
    \centering
    \includegraphics[width=\textwidth]{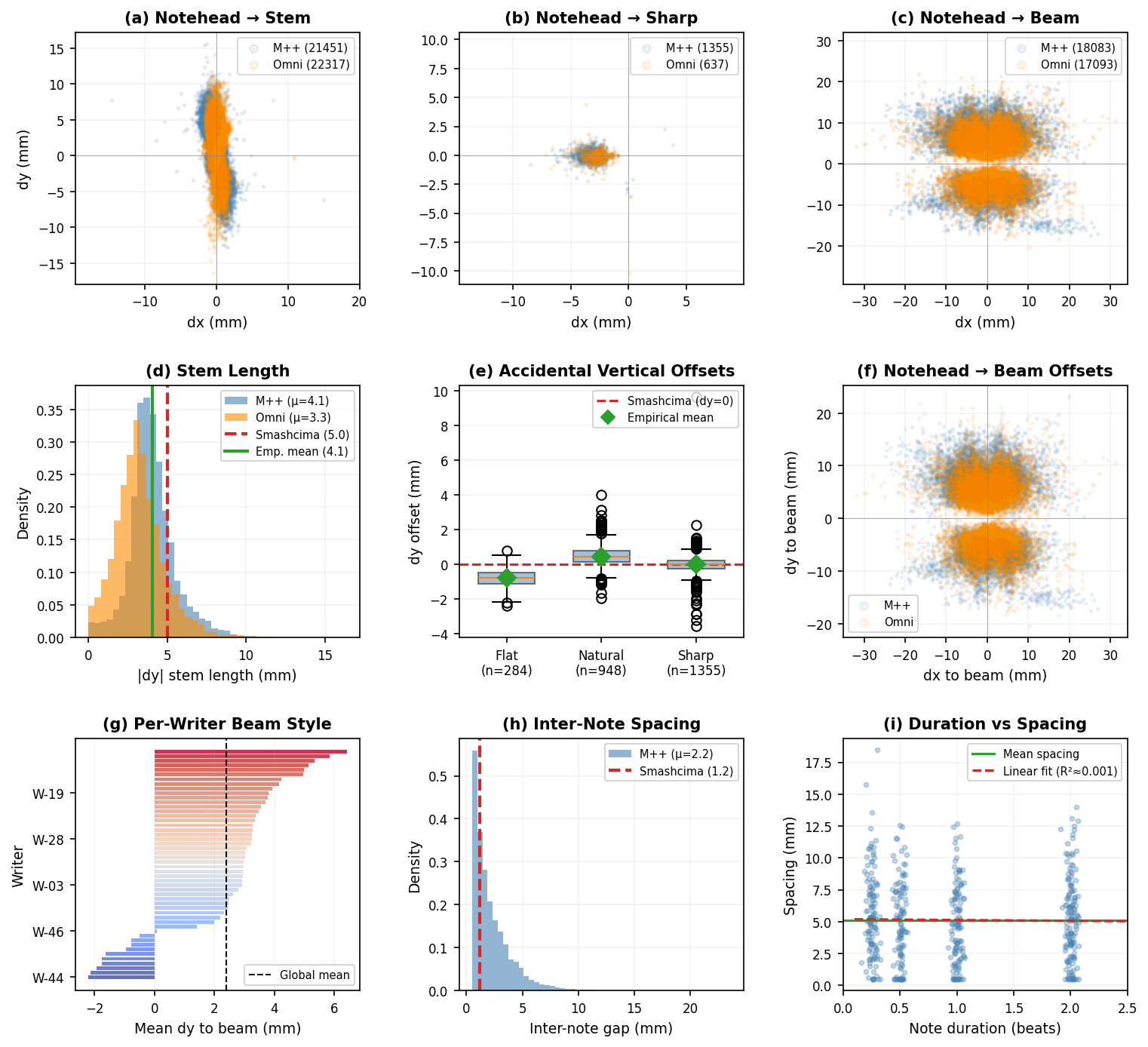}
    \caption{Empirical positioning distributions. Row 1: cross-dataset scatterplots (blue: MUSCIMA++, orange: OmniOMR). Row 2: key findings with Smashcima baselines (red dashed) and empirical means (green). Row 3: writer-dependent beam style, inter-note spacing, and the absence of duration-proportional spacing ($\text{R}^2 \approx 0.001$).}
    \label{fig:empirical_overview}
\end{figure*}

Figure~\ref{fig:empirical_overview} summarises the key empirical findings across both datasets. We detail each relationship below.

Smashcima uses a fixed stem length of \SI{5.0}{mm}. The empirical distributions are direction-conditional: stem-up mean $= \SI{3.8}{mm}$ ($\sigma = 1.5$), stem-down mean $= \SI{4.2}{mm}$ ($\sigma = 1.5$). Replacing the constant with the empirical mean reduces RMSE by 57--60\% (Table~\ref{tab:results}). Per-writer models perform identically to the global model under LOWO, suggesting between-writer variance is small relative to within-writer noise. Cross-dataset validation shows that OmniOMR stem lengths are slightly shorter (\SI{3.4}{mm} up, \SI{3.3}{mm} down) but the direction is consistent.

Smashcima uses $dy = 0$ for all accidentals. Our analysis shows flats sit at $dy = \SI{-0.78}{mm}$ (above centre), naturals at $dy = \SI{+0.45}{mm}$ (below centre), and sharps at $dy \approx 0$, consistent across all 50 writers, producing 9--24\% RMSE improvement and per-writer conditioning provides no benefit. These offsets are measured between bounding box centres, the non-zero values for flat and natural reflect the asymmetric shape of these glyphs (e.g., the flat's tall vertical stroke shifts its bbox centre above the visual attachment point).

Beam angle depends on pitch contour and writer style. The linear model $\theta = \SI{4.0}{\degree\per\milli\metre} \times \text{pitch\_proxy} + \epsilon$ achieves $\text{R}^2 = 0.59$ in MUSCIMA++ and $\text{R}^2 = 0.66$ in OmniOMR. This is the only relationship where writer conditioning provides meaningful improvement (+9\% RMSE reduction).

Smashcima uses a fixed \SI{1.2}{mm} gap between note columns, the empirical mean is \SI{5.07}{mm} in MUSCIMA++ and \SI{10.88}{mm} in OmniOMR. This suggests that duration does not predict spacing ($\text{R}^2 = 0.001$), and no positional effect was seen across the system width.

\subsection{Cross-Dataset Comparison}
\label{sec:crossdataset}

The 2$\times$ spacing difference found between MUSCIMA++ and OmniOMR is because of the density effect, while MUSCIMA++ contains 62.9 noteheads per page (0.28/mm), OmniOMR staves have 12.0 (0.085/mm). This means the spacing parameter depends on target note density, not dataset of origin.

\begin{table}[t]
    \centering
    \caption{Summary of positioning corrections. RMSE improvement is relative to Smashcima's hard-coded baseline, evaluated under LOWO cross-validation.}
    \label{tab:results}
    \begin{tabular}{@{}lcccc@{}}
        \toprule
        \textbf{Relationship} & \textbf{Baseline} & \textbf{Empirical} & \textbf{RMSE} & \textbf{Improv.} \\
         & \textbf{value} & \textbf{mean} & \textbf{(global)} & \\
        \midrule
        Stem length (up) & \SI{5.0}{mm} & \SI{3.8}{mm} & \SI{1.44}{mm} & +60\% \\
        Stem length (down) & \SI{5.0}{mm} & \SI{4.2}{mm} & \SI{1.32}{mm} & +57\% \\
        Beam angle (global) & \SI{0}{\degree} & \SI{4.0}{\degree/mm} & \SI{4.51}{\degree} & +40\% \\
        Beam angle (per-writer) & \SI{0}{\degree} & writer int. & \SI{3.81}{\degree} & +49\% \\
        Inter-note gap & \SI{1.2}{mm} & \SI{5.07}{mm} & \SI{3.94}{mm} & +29\% \\
        Accidental flat ($dy$) & \SI{0}{mm} & \SI{-0.78}{mm} & \SI{0.96}{mm} & +24\% \\
        Accidental natural ($dy$) & \SI{0}{mm} & \SI{+0.45}{mm} & \SI{0.91}{mm} & +9\% \\
        Accidental sharp ($dx$) & \SI{-3.5}{mm} & \SI{-3.25}{mm} & \SI{0.98}{mm} & +4\% \\
        \bottomrule
    \end{tabular}
\end{table}

\section{GNN for Context-Dependent Positioning}
\label{sec:gnn}

The empirical mean approach treats each edge independently. However, glyph positions are not independent: stem length depends on beam group size, beam angle constrains all attached stems jointly, and crowded measures compress offsets. We train a Graph Neural Network on the MuNG graph structure to predict context-dependent $(dx, dy)$ offsets.

\subsection{Architecture}

Each MuNG page is converted to a PyTorch Geometric graph where nodes represent glyphs and directed edges represent attachment relationships. Node features include a learned class embedding (32-dim), bounding box dimensions (width, height), and spatial position. Edge features include a learned edge-type embedding (16-dim).

The model consists of a GraphSAGE~\cite{graphsage} encoder (4 message-passing layers, 128-dim hidden, residual connections, layer normalisation) followed by an edge decoder MLP. For each directed edge, the decoder takes the concatenation of source embedding, target embedding, their element-wise product, their difference, and the edge-type embedding, and predicts $(dx, dy)$ in millimetres. The model has 236,610 trainable parameters and is trained with Huber loss ($\delta = \SI{1.0}{mm}$) using AdamW with cosine annealing.

\subsection{Results}

Under LOWO cross-validation (50 folds), the GNN achieves an overall RMSE of $2.06 \pm 0.30$~mm across all edge types. The per-edge-type breakdown (Table~\ref{tab:gnn_results}) demonstrates that the GNN's advantage is focused on context-dependent relationships:

\begin{table}[t]
    \centering
    \caption{Per-edge-type positioning RMSE (mm). The GNN improves most on relationships where placement is constrained by local context.}
    \label{tab:gnn_results}
    \begin{tabular}{@{}lccccc@{}}
        \toprule
        \textbf{Edge type} & \textbf{Baseline} & \textbf{Empirical} & \textbf{GNN} & \textbf{vs Emp.} & $n$ \\
        \midrule
        notehead $\to$ stem & 3.07 & 1.44 & \textbf{0.60} & +80.6\% & 1984 \\
        notehead $\to$ beam & 6.89 & 7.13 & \textbf{1.07} & +84.4\% & 1743 \\
        notehead $\to$ flag & 3.68 & 3.68 & \textbf{0.75} & +79.5\% & 323 \\
        notehead $\to$ ledger line & 1.16 & 1.16 & \textbf{0.55} & +52.7\% & 600 \\
        notehead $\to$ staccato & 2.59 & 2.59 & \textbf{0.91} & +65.0\% & 289 \\
        notehead $\to$ duration dot & 0.62 & 0.62 & \textbf{0.61} & +1.5\% & 131 \\
        notehead $\to$ natural & 0.53 & 0.53 & 0.56 & $-$6.5\% & 97 \\
        \bottomrule
    \end{tabular}
\end{table}

\begin{figure}[t]
    \centering
    \includegraphics[width=0.9\textwidth]{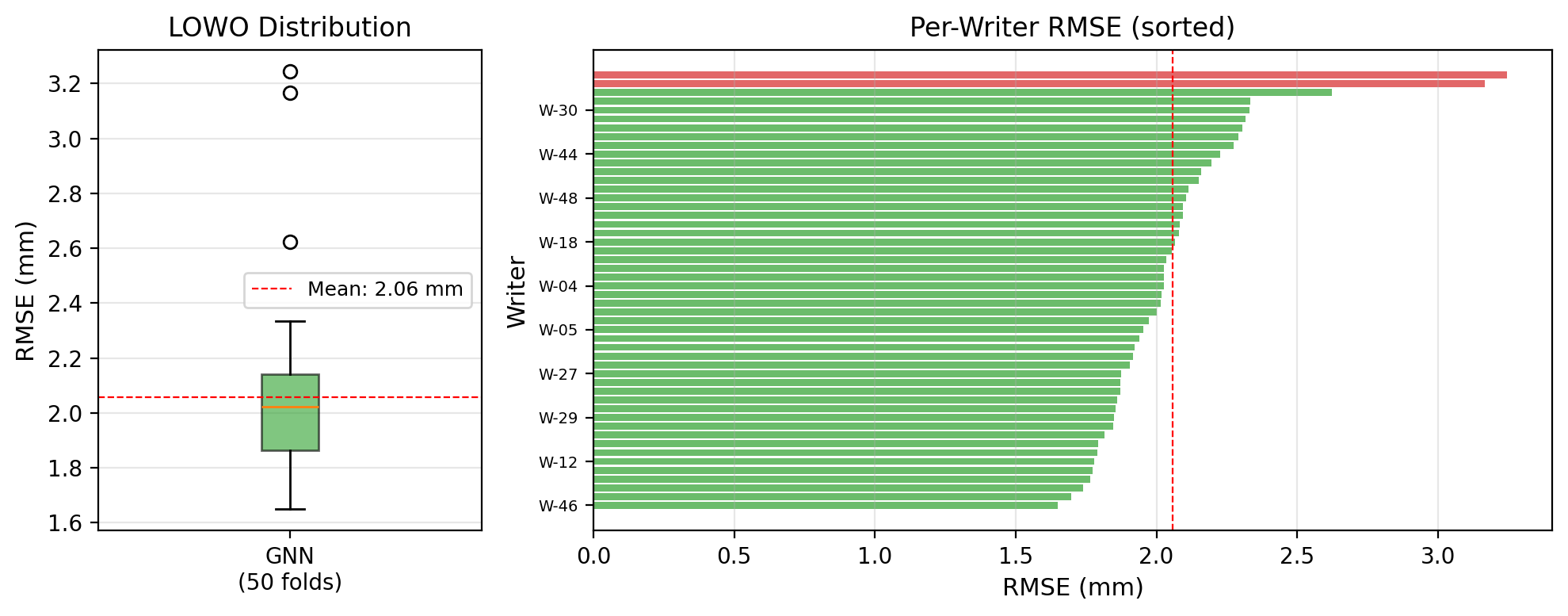}
    \caption{LOWO cross-validation: per-writer RMSE distribution across 50 held-out writers. The GNN generalises consistently (mean $2.06 \pm 0.30$~mm), with only two outliers above 3~mm.}
    \label{fig:lowo}
\end{figure}

The GNN achieves 50--85\% improvement over empirical means on stems, beams, and flags, specifically the relationships where neighbouring glyphs constrain placement (e.g., stem length depends on beam group size). On tight, context-independent relationships (accidentals, duration dots), the GNN matches the empirical mean, showing it does not degrade performance where context is uninformative. Two outlier writers (W-09: 3.24~mm, W-20: 3.17~mm) have unusual styles but the model still generalises reasonably (Figure~\ref{fig:lowo}).

\section{Downstream Evaluation}
\label{sec:evaluation}

To evaluate whether improved positioning translates to better OMR performance, we train object detection models on synthetic data generated by three synthesiser variants: baseline (Smashcima hard-coded), empirical (data-driven means), and GNN (context-dependent positioning).

\subsection{Experimental Setup}

We generate 1000 pages of synthetic handwritten music from 100 MusicXML sources using each synthesiser variant, with identical random seeds and glyph sampling. The only difference between datasets is the positioning of symbols. Bounding box annotations are extracted from the synthesis scene graph in COCO format, covering 20 symbol categories. We train YOLOv8-nano~\cite{yolov8} on each dataset for 50 epochs with identical hyperparameters (batch size 16, image size 640, AdamW optimiser).

The GNN-positioned synthesiser uses the learned stem length distributions from the GNN, context-dependent means (single-note stems: \SI{3.8}{mm}; beamed stems: \SI{2.3}{mm}, adjusted by group size) with calibrated variance from the GNN's residual distribution.

\begin{figure}[t]
    \centering
    \includegraphics[width=0.32\textwidth]{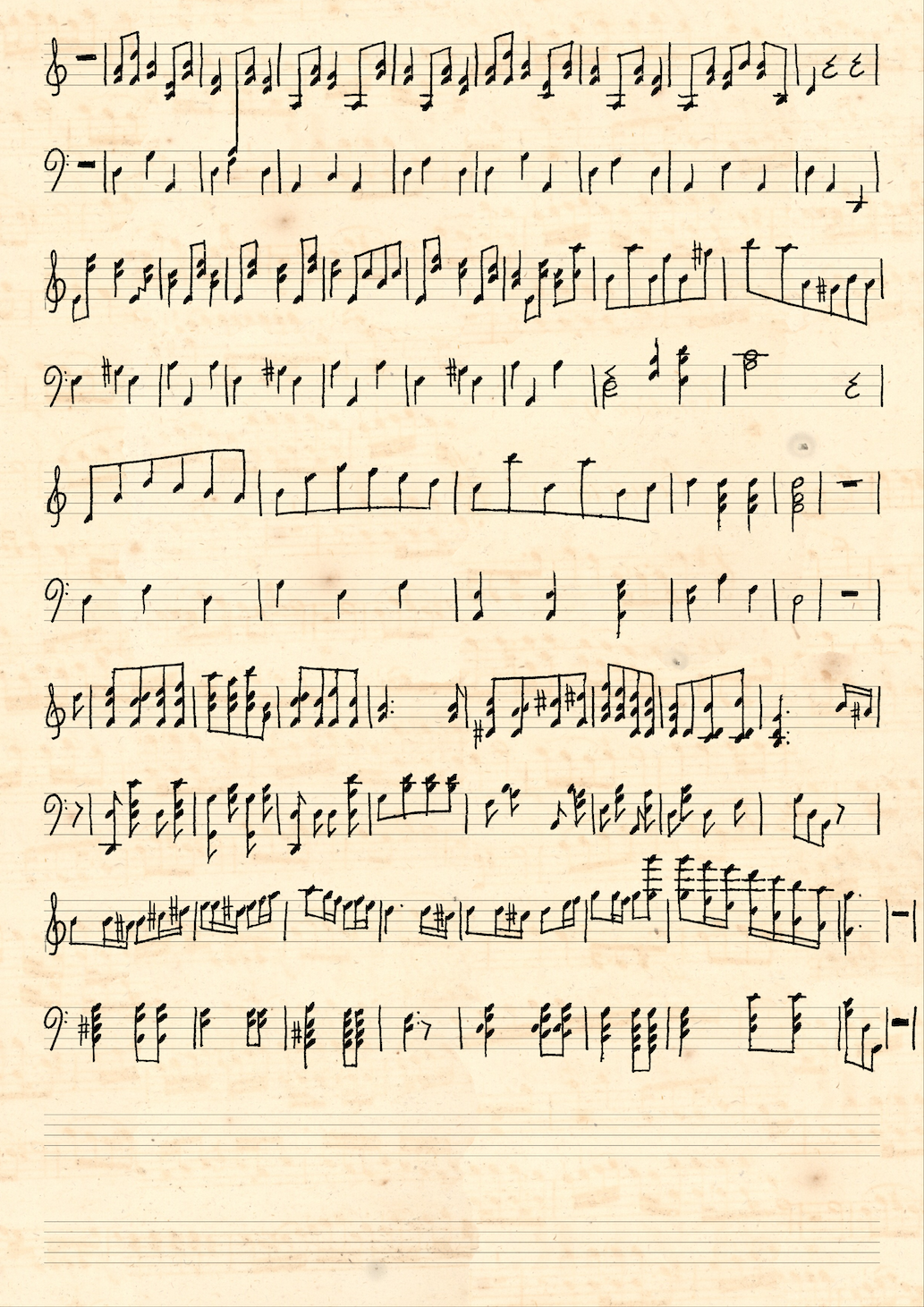}
    \hfill
    \includegraphics[width=0.32\textwidth]{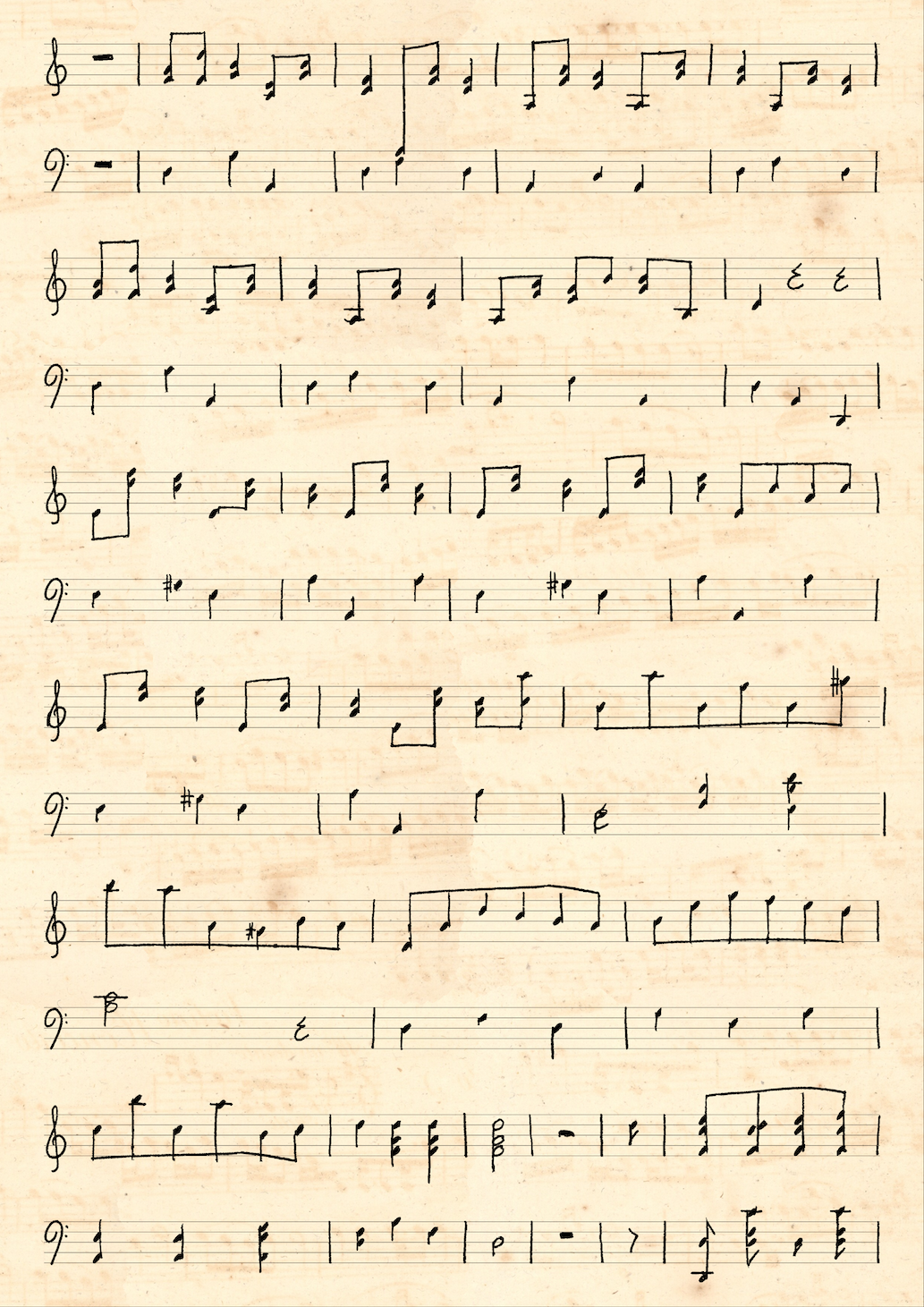}
    \hfill
    \includegraphics[width=0.32\textwidth]{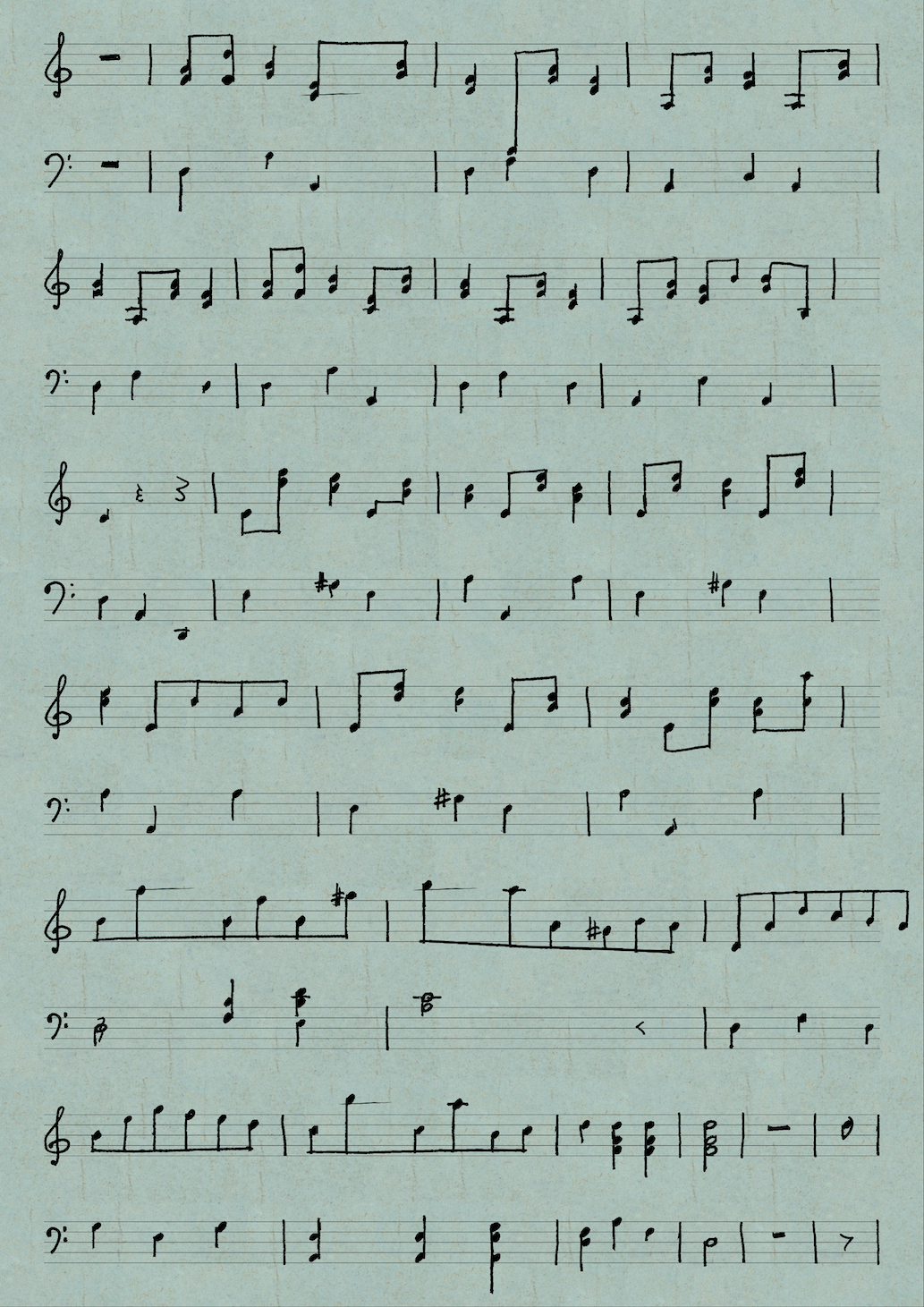}
    \caption{Synthesised output from the same MusicXML input and random seed. Left: baseline (hard-coded constants). Centre: empirical (data-driven means). Right: GNN (context-dependent positioning). Note the progressively shorter stems, tilted beams, and wider inter-note spacing from left to right. The GNN output shows more realistic stem length variation in beamed groups.}
    \label{fig:comparison}
\end{figure}

\subsection{Synthetic Validation}

On synthetic validation data, the three models achieve mAP50 of 0.297 (baseline), 0.444 (empirical, +50\%), and 0.520 (GNN, +75\% over baseline, +17\% over empirical). The GNN model's improvement is consistent across symbol classes, with the largest gains on stems and beams.

\subsection{Transfer to Real Handwritten Scores}

To evaluate whether the synthetic pretraining advantage transfers to real data, we conduct a few-shot domain adaptation experiment. All three pretrained models are fine-tuned on varying amounts of real OmniOMR stave images (5--924) and evaluated on a fixed held-out set of 231 real staves. Each condition is repeated with 3 random subsets.

\begin{table}[t]
    \centering
    \caption{Few-shot fine-tuning on real handwritten staves (mAP50). The GNN-pretrained model consistently outperforms both baselines across all data regimes.}
    \label{tab:fewshot}
    \begin{tabular}{@{}rccccc@{}}
        \toprule
        \textbf{Shots} & \textbf{Baseline} & \textbf{Empirical} & \textbf{GNN} & \textbf{$\Delta$ vs Base} & \textbf{$\Delta$ vs Emp} \\
        \midrule
        5 & 0.093 & 0.101 & \textbf{0.114} & +22.6\% & +12.9\% \\
        10 & 0.104 & 0.106 & \textbf{0.125} & +20.8\% & +18.3\% \\
        20 & 0.110 & 0.109 & \textbf{0.126} & +15.1\% & +16.1\% \\
        50 & 0.115 & 0.116 & \textbf{0.136} & +18.7\% & +17.5\% \\
        100 & 0.123 & 0.137 & \textbf{0.153} & +24.0\% & +11.4\% \\
        200 & 0.168 & 0.196 & \textbf{0.218} & +30.0\% & +11.3\% \\
        500 & 0.263 & 0.271 & \textbf{0.294} & +12.0\% & +8.6\% \\
        924 & 0.323 & 0.323 & \textbf{0.345} & +6.7\% & +6.7\% \\
        \bottomrule
    \end{tabular}
\end{table}

\begin{figure}[t]
    \centering
    \includegraphics[width=0.9\textwidth]{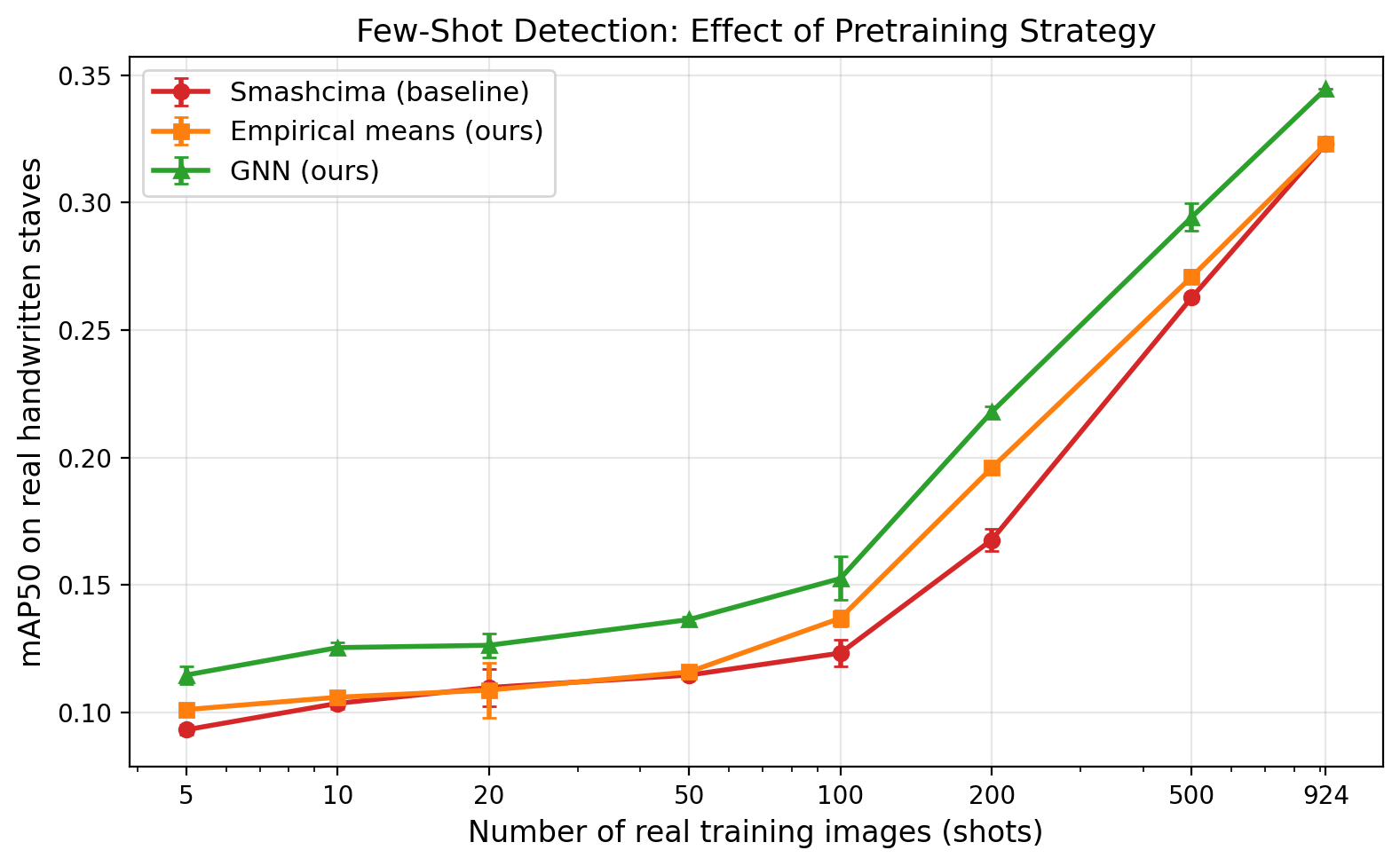}
    \caption{Few-shot domain adaptation on real OmniOMR staves. The GNN-pretrained model (green) consistently outperforms both the empirical (orange) and baseline (red) models across all data regimes. The advantage is largest in the low-data regime where pretraining quality matters most.}
    \label{fig:fewshot}
\end{figure}

The results (Table~\ref{tab:fewshot}, Figure~\ref{fig:fewshot}) show that the GNN-pretrained model provides a consistent advantage across all data regimes. At 200 real training images, the GNN model achieves mAP50 of 0.218 compared to 0.196 for empirical (+11.3\%) and 0.168 for baseline (+30.0\%). The advantage persists even at 924 images (+6.7\%), suggesting the quality difference in synthetic data is real, not a side effect of limited annotations.

\section{Discussion}
\label{sec:discussion}

The GNN improves most on relationships where local graph structure provides information beyond the marginal distribution. Stem length in isolation has high variance (\SI{1.5}{mm} std), but conditioned on beam group membership, the variance drops: beamed stems are shorter (\SI{2.3}{mm}) than single-note stems (\SI{3.8}{mm}). The GNN learns this through message passing, without feature engineering.

The empirical mean approach cannot capture this because it computes one mean per edge type regardless of context. The GNN sees the 4-hop neighbourhood (4 message-passing layers) and can distinguish a stem connected to a 4-note beam from one connected to a 2-note beam.

\subsection{The Layout Ceiling}

Going from hard-coded constants to empirical means to GNN shows the structure of the layout component of the going from synthetic to real gap. Without any machine learning we achieve +50\% mAP50 on synthetic data. Adding context-dependent positioning via the GNN yields a further +17\% on synthetic and +11\% on real data. This suggests that layout fidelity deserves at least as much attention as model architecture or symbol appearance in synthesis-based OMR, though this will need to be measured also in the end-to-end context.

Across all relationships studied, only beam angle benefits from writer-specific models. For accidentals and stem length, per-writer models perform identically to the global model under LOWO evaluation. A single global model is sufficient for most positioning parameters when synthesising for unseen writers.

The 2$\times$ spacing difference between MUSCIMA++ and OmniOMR is a density effect rather than a stylistic one. MUSCIMA++ writers copied dense music onto fixed-width paper, forcing compression. The appropriate spacing for synthesis depends on target note density, not dataset of origin. Duration-proportional spacing, standard in engraving software, does not hold in handwriting ($\text{R}^2 = 0.001$), so systems targeting handwritten output should not use it.

\subsection{Limitations}

Our GNN integration uses the learned distributions at synthesis time rather than full graph inference, which would require the graph to exist before positioning. A tighter integration that iteratively refines positions using the GNN on the constructed scene graph is left for future work.

The detection model operates at 640px resolution, so small symbols (augmentation dots, ledger lines) are difficult to detect regardless of positioning quality. The 20-class mAP50 averages over both easy categories (clefs: 0.99) and hard ones (dots: 0.0).

An interesting next step would be to compare the effect of layout corrections against using more realistic symbol appearances (e.g., OmniOMR glyphs or GAN-generated symbols~\cite{gasbert2024}), to further decompose the factors contributing to the synthetic-to-real domain gap.

\section{Conclusion}
\label{sec:conclusion}

We have presented a data-driven approach to glyph positioning in handwritten music synthesis, going from empirical characterisation to learned context-dependent prediction. Replacing Smashcima's hard-coded constants with empirical means reduces positioning RMSE by 4--60\% with no model training. A GNN trained on the MuNG graph structure achieves an additional 50--85\% reduction on context-dependent relationships (stems, beams, flags) by modelling how neighbouring glyphs constrain placement.

This also improves downstream detection, a YOLOv8 detector pretrained on GNN-positioned synthetic data achieves +30\% mAP50 over the baseline and +11\% over empirical means when fine-tuned on 200 real handwritten staves. The advantage persists across all few-shot regimes (5--924 images), showing that positioning accuracy in synthesis directly translates to better OMR performance.

These results show that layout fidelity is an important component of the synthetic-to-real domain gap in OMR. Each level captures information the previous cannot, and each translates to measurable improvement in downstream OMR.


\begin{credits}
\subsubsection{\ackname}
This work has been supported by the Ministry of Culture of the Czech Republic (project OmniOMR of the NAKI III programme, no. DH23P03OVV008), the SVV project no. 260 821, and by the project “Human-centred AI for a Sustainable and Adaptive Society” (reg. no. CZ.02.01.01/00/23\_025/0008691), co-funded by the European Union. The computing infrastructure was provided by the LINDAT/CLARIAH-CZ Research Infrastructure,\footnote{\url{https://lindat.cz}} supported by the Ministry of Education, Youth and Sports of the Czech Republic (project no. LM2023062).

\end{credits}


\bibliographystyle{splncs04}
\bibliography{references}

\end{document}